\documentclass[letterpaper]{article} 
\usepackage{aaai2026}  
\usepackage{times}  
\usepackage{helvet}  
\usepackage{courier}  
\usepackage[hyphens]{url}  
\usepackage{graphicx} 
\usepackage{natbib}  
\usepackage{caption} 
\usepackage{algorithm}
\usepackage{algorithmic}
\usepackage{amsmath}
\usepackage{amsfonts}
\usepackage{amssymb}
\usepackage{graphicx}
\usepackage{booktabs}

\usepackage{newfloat}
\usepackage{listings}
\DeclareCaptionStyle{ruled}{labelfont=normalfont,labelsep=colon,strut=off} 
\floatstyle{ruled}
\newfloat{listing}{tb}{lst}{}
\floatname{listing}{Listing}
\title{Reinforcement Learning enhanced Online Adaptive Clinical Decision Support via Digital Twin powered Policy and Treatment Effect optimized Reward}
\author {
    Xinyu Qin\textsuperscript{\rm 1}, Ruiheng Yu\textsuperscript{\rm 1},
    Lu Wang\textsuperscript{\rm 1,\rm 2}\thanks{Corresponding author}
}
\affiliations {
    \textsuperscript{\rm 1}Department of Biomedical Engineering, Cullen College of Engineering, University of Houston\\
    \textsuperscript{\rm 2}Department of Health Systems \& Population Health Sciences, Tilman J. Fertitta Family College of Medicine, University of Houston\\
    xqin5@cougarnet.uh.edu, ryu11@cougarnet.uh.edu, lwang71@central.uh.edu
}

\usepackage{bibentry}

\begin{document}

\maketitle

\begin{abstract}
Clinical decision support must adapt online under safety constraints. We present an online adaptive tool where reinforcement learning provides the policy, a patient digital twin provides the environment, and treatment effect defines the reward. The system initializes a batch-constrained policy from retrospective data and then runs a streaming loop that selects actions, checks safety, and queries experts only when uncertainty is high. Uncertainty comes from a compact ensemble of five Q-networks via the coefficient of variation of action values with a $\tanh$ compression. The digital twin updates the patient state with a bounded residual rule. The outcome model estimates immediate clinical effect, and the reward is the treatment effect relative to a conservative reference with a fixed z-score normalization from the training split. Online updates operate on recent data with short runs and exponential moving averages. A rule-based safety gate enforces vital ranges and contraindications before any action is applied. Experiments in a synthetic clinical simulator show low latency, stable throughput, a low expert query rate at fixed safety, and improved return against standard value-based baselines. The design turns an offline policy into a continuous, clinician-supervised system with clear controls and fast adaptation.
\end{abstract}

\section{Introduction}
Clinical decisions arrive in sequence and involve risk \citep{SuttonBarto2018}. Policies learned offline can be effective at deployment, yet dataset shift and limited coverage reduce value as conditions evolve \citep{Levine2020OfflineRL,Jayaraman2024PrimerRLMedicine}. Our objective is an online adaptive clinical decision support tool that learns during use while respecting safety. Reinforcement learning (RL) drives long-horizon optimization with explicit value and policy models \citep{SuttonBarto2018,Levine2020OfflineRL}. A patient digital twin (DT) provides an executable environment that supports state updates and short rollouts consistent with the incoming stream \citep{Viceconti2021DigitalTwinHC}. Treatment effect (TE) defines the reward so that learning aligns with clinical benefit under a clear counterfactual reference \citep{HernanRobins2020}.

We link these parts into a single system focused on online learning with guardrails. First, an offline stage trains a batch-constrained policy on retrospective data to respect dataset support \citep{Fujimoto2019BCQ}. Second, a streaming loop selects actions by the mean of a compact Q-ensemble, checks safety with a rule-based gate, and queries experts only when uncertainty is high. Uncertainty comes from the coefficient of variation of action values across five heads with a $\tanh$ squashing, which is a simple and reliable uncertainty quantification strategy for deep ensembles \citep{Lakshminarayanan2017DeepEnsembles}. Third, short online updates adjust models on recent labeled data and use exponential moving averages to preserve stability under drift \citep{Jayaraman2024PrimerRLMedicine}. Label effort is controlled by a $k$-center selection that favors diverse high-uncertainty cases \citep{Sener2018CoreSet}. The tool exposes small controls for the query threshold, stream rate, and batch size so behavior can change without full retraining.

This paper makes three technical contributions that integrate RL, DT, and TE into an online adaptive tool for decision support. It places RL at the core of clinical decision making with a stable offline initializer and a lightweight online loop. It embeds a DT that enables fast and consistent state updates during streaming operation. It shapes learning with a TE-based reward so that the policy improves outcomes that matter clinically while an uncertainty rule and a safety gate limit risk. Results in a synthetic clinical setting show improved return and efficiency at fixed safety compared with common value-based baselines.

\begin{itemize}
    \item \textbf{A safety-aware online evaluation loop for DT4H.} We integrate an uncertainty-driven query mechanism with explicit rule-based safety gates (vital-sign plausibility, medication dose bounds, conflict checks with blacklisted co-medications, and data-quality screens) to trigger conservative fallbacks before any potential violation.
    \item \textbf{Label-efficient active learning under strict latency.} We formalize online querying using the coefficient of variation of Q-ensemble action values with a $\tanh$ compression and show a low query rate at sustained throughput, consistent with evidence that deep ensembles provide reliable uncertainty under resource budgets \citep{Lakshminarayanan2017DeepEnsembles,CheaperEnsembles2024}.
    \item \textbf{Seamless offline-to-online adaptation.} We initialize from competitive offline baselines and perform frequent small updates with exponential moving averages, which balances plasticity and stability in nonstationary streams \citep{Fujimoto2019BCQ,Jayaraman2024PrimerRLMedicine}.
    \item \textbf{Human-centered oversight via LLMs.} Medical LLMs are used for rationale surfacing and documentation only, acknowledging their strengths and current limitations in clinical deployment \citep{Singhal2023LLMClinical,Hager2024LLMLimitations}.
    \item \textbf{Framework-agnostic de-identification at data ingress.} We add a policy-driven module that removes direct identifiers, pseudonymizes keys, generalizes quasi-identifiers (e.g., ZIP$\rightarrow$ZIP3 and age bucketing), and applies bounded date shifting. $k$-anonymity checks with logs ensure coverage, and only de-identified data enter the pipeline (HIPAA Safe Harbor).
\end{itemize}

\section{Methodology}
\begin{figure}[t]
  \centering
  \includegraphics[width=\linewidth]{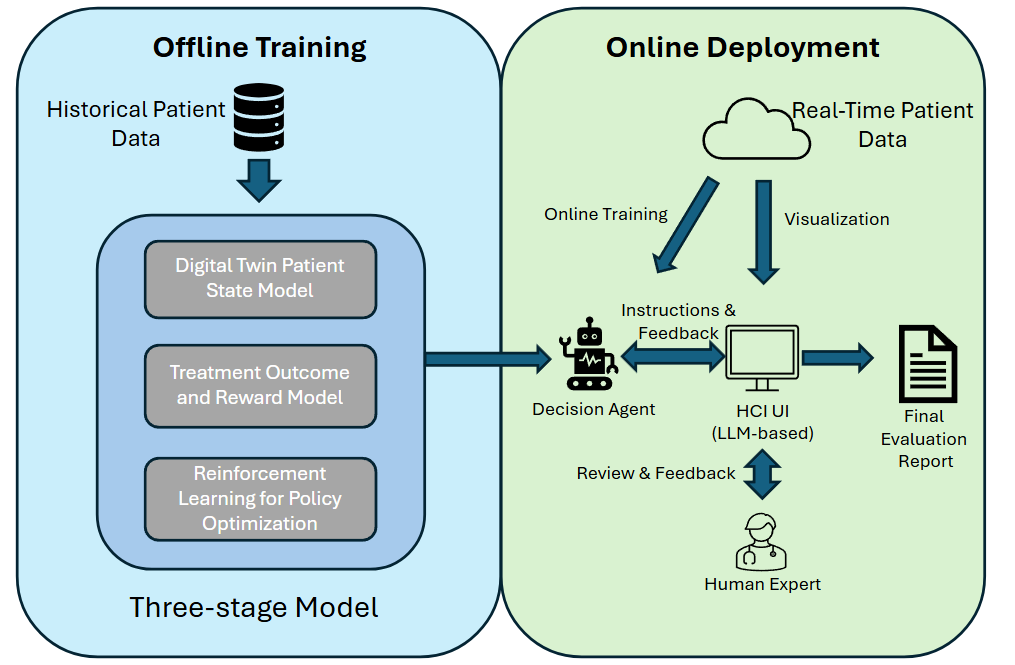}
  \caption{Overview of the proposed offline-to-online framework.}
  \label{fig:framework}
\end{figure}

\subsection{Overview}

The framework consists of four interconnected components that work synergistically to provide continuous learning capabilities: (1) an offline-trained base model that provides initial clinical knowledge, (2) an online learning system with active sampling, (3) an LLM-based interaction layer for interpretability, and (4) a human-computer interface (HCI) designed for clinical workflows. An framework overview is shown in Figure~\ref{fig:framework}.

\subsection{Offline Training}

Before any model consumes data, we run a policy-driven de-identification pass to ensures that all learning and evaluation operate on \emph{de-identified} data consistent with Health Insurance Portability and Accountability Act (HIPAA) standards. The detailed operations can be viewed in supplementary material due to page limits.

\subsubsection*{Stage 1: Dynamics Model (Ensemble of Five)}
We construct a patient digital twin that predicts the next state from recent history and the applied treatment. 
The model is a Transformer encoder that receives a sequence of state vectors and the aligned action tokens, with a causal attention mask and a padding mask. 
At each step the network predicts a residual change and we apply a bounded update to improve stability during multi-step rollouts:
\begin{equation}
\mathbf{s}_{t+1} \;=\; \operatorname{clip}\!\Bigl(\mathbf{s}_{t} \,+\, 0.05\, \tanh\!\bigl(f_{\theta}(\mathbf{s}_{0:t}, a_{0:t})\bigr),\; 0,\; 1\Bigr).
\end{equation}

Here $\mathbf{s}_t \in [0,1]^d$ is the normalized state and $a_t \in \{0,\dots,K-1\}$ is the discrete action. 
The loss is computed only on valid timesteps within each sequence by a binary mask that ignores padding. 
We use a Smooth L1 objective over one step predictions:
\begin{equation}
\mathcal{L}_{\text{DT}}(\theta) \;=\; \frac{1}{|\Omega|} \sum_{(i,t)\in \Omega} 
\ell_{\text{smooth}}\!\left(\hat{\mathbf{s}}^{(i)}_{t+1} ,\, \mathbf{s}^{(i)}_{t+1}\right),
\end{equation}

where $\Omega$ denotes all valid positions in the mini batch. 
Training uses AdamW, gradient clipping, and a learning rate scheduler. 
We train five independent models under different seeds and keep all five for evaluation. 
During rollout we aggregate the predictions by the ensemble mean. 
We also use the ensemble variance as an uncertainty signal.

\subsubsection*{Stage 2: Counterfactual Treatment Outcome and Reward Model}
The outcome model $r_{\phi}$ estimates the immediate outcome signal from the current state and the action. 
The network encodes the state into a health representation and embeds the action, then concatenates both and passes them to a small prediction head that outputs a scalar. 
To reduce spurious treatment information in the health representation we add an adversarial penalty with a treatment discriminator $D_{\xi}(a\,|\,\mathbf{z}_{\text{health}})$. 
The objective is
\begin{equation}
\small
\min_{\phi}\;\max_{\xi}\;\;\mathbb{E}_{(\mathbf{s},a,y)\sim\mathcal{D}}\!\left[\,|\,r_{\phi}(\mathbf{s},a)-y\,| 
\;+\; \lambda\, \mathcal{L}_{\text{adv}}\!\bigl(D_{\xi}(a\,|\,\mathbf{z}_{\text{health}})\bigr)\right],
\end{equation}

where $\lambda>0$ balances predictive accuracy and the adversarial term. 
We compute reward normalization statistics $(\mu_r,\sigma_r)$ on the training set and persist them. 
All downstream components use the normalized reward $\tilde r=(r-\mu_r)/\sigma_r$.

\subsubsection*{Stage 3: Offline Policy Learning with BCQ}
We adopt Batch Constrained Q learning for discrete actions. 
BCQ restricts action choices to those that are likely under the behavior policy observed in the dataset and then selects the action with the highest estimated value. 
We implement the discrete variant with a dueling Q network. 
Given the dataset $\mathcal{D}$ and the normalized reward, BCQ learns $Q_{\psi}$ and a behavior model $b(a\,|\,\mathbf{s})$. 
The policy acts greedily over the constrained set
\begin{equation}
\small
\begin{aligned}
\pi(\mathbf{s}) \;&=\; \arg\max_{a \in \mathcal{A}_{\text{valid}}(\mathbf{s})} Q_{\psi}(\mathbf{s},a), \\[6pt]
\mathcal{A}_{\text{valid}}(\mathbf{s}) \;&=\; \{\,a \in \mathcal{A}\;:\; b(a\,|\,\mathbf{s}) \ge \tau_{\text{supp}}\,\}.
\end{aligned}
\end{equation}

We select $\tau$ through validation. 
We train BCQ with the same replay data used by the dynamics and outcome stages. 
The best policy checkpoint is saved in a portable format and used during evaluation.

\subsection{Online Learning with Active Sampling}

The transition from offline to online learning presents unique challenges in clinical settings where incorrect decisions can have serious consequences. We address these challenges through a carefully designed online learning pipeline that maintains safety while enabling continuous adaptation. High-uncertainty candidates (\( \tilde{u}(s_t)>\tau \)) are first buffered; once the buffer reaches \(k\) items, we apply a \(k\)-center selection (uncertainty-weighted farthest-first) to query a \emph{batch} of diverse samples in one shot, otherwise we keep buffering until reaching \(k\).

\subsubsection{Uncertainty-Based Active Learning}
We maintain an ensemble of $H=5$ Q-networks $\{Q_{\psi_k}\}_{k=1}^{H}$ carried over from the offline stage to quantify epistemic uncertainty in treatment decisions.
At test time, the action is chosen greedily with respect to the ensemble mean:
\begin{equation}
a_t \;=\; \arg\max_{a \in \mathcal{A}} \; \frac{1}{H}\sum_{k=1}^{H} Q_{\psi_k}(s_t,a).
\end{equation}

For each state--action pair we compute the ensemble mean and standard deviation:
\begin{equation}
\mu_a(s_t) \;=\; \frac{1}{H}\sum_{k=1}^{H} Q_{\psi_k}(s_t,a),
\end{equation}
\begin{equation}
\sigma_a(s_t) \;=\; \sqrt{\frac{1}{H-1}\sum_{k=1}^{H}\!\bigl(Q_{\psi_k}(s_t,a)-\mu_a(s_t)\bigr)^2}.
\end{equation}
We then form the coefficient of variation:
\begin{equation}
\mathrm{CV}_a(s_t) \;=\; \frac{\sigma_a(s_t)}{|\mu_a(s_t)|+\epsilon}.
\end{equation}
Our decision statistic is the $\tanh$-squashed maximum across actions:
\begin{equation}
\tilde{u}(s_t) \;=\; \tanh\!\Bigl(\max_{a \in \mathcal{A}} \mathrm{CV}_a(s_t)\Bigr).
\label{eq:uncertainty-stat}
\end{equation}
We query an expert label iff $\tilde{u}(s_t) > \tau$; unless noted, $\tau=0.2$ for all methods. Queried samples are appended to the labeled buffer and can trigger online updates.

When a BCQ policy is active (no ensemble heads), we replace \eqref{eq:uncertainty-stat} with a normalized state-variance proxy $\hat{u}(s_t)\!\in\![0,1]$ and apply the same threshold~$\tau$.

When multiple high-uncertainty samples accumulate, we select a size-$k$ batch by a $k$-center objective to promote state-space coverage. Let $\mathcal{U}$ be the pool of candidates exceeding the threshold and let $d(\cdot,\cdot)$ denote Euclidean distance in state space. We choose:
\begin{equation}
\operatorname*{selected}
\;=\;
\arg\max_{\mathcal{B}\subseteq\mathcal{U},\,|\mathcal{B}|=k}
\;\min_{\,\mathbf{s}\in \mathcal{U}\setminus \mathcal{B}}
\;\max_{\,\mathbf{s}'\in \mathcal{B}}
d(\mathbf{s},\mathbf{s}') \cdot \tilde{u}(\mathbf{s}).
\end{equation}
This favors diverse and informative states while keeping the querying budget small.

\subsubsection{Incremental Model Updates}

Instead of retraining entire models, we implement targeted incremental updates that preserve learned knowledge while incorporating new information. For the Transformer dynamics model $\hat{f}_\theta$ with layers $\{l_1, ..., l_n\}$, we freeze parameters $\theta_{1:n-2}$ and update only $\theta_{n-1:n}$:

\begin{equation}
\theta_{t+1}^{(n-1:n)} \;=\; 
\theta_t^{(n-1:n)} \;-\; \eta \nabla_{\theta_{n-1:n}} 
\mathcal{L}(\theta_t;\, \mathcal{D}_t^{\text{new}}).
\end{equation}

To maintain stability during online updates, we employ exponential moving averages for critical parameters:

\begin{equation}
\bar{\theta}_{t+1} \;=\; \alpha \bar{\theta}_t + (1-\alpha)\theta_{t+1}, 
\quad \alpha = 0.99.
\end{equation}

This EMA mechanism provides a crucial balance between adaptability to new patterns and retention of previously learned knowledge.

\subsubsection{Experience Replay with Prioritization}

We maintain two experience buffers: a labeled buffer $\mathcal{B}_L$ (capacity 10K) for expert-validated transitions and a weak buffer $\mathcal{B}_W$ (capacity 50K) for model-predicted labels. Sampling prioritizes recent labeled data while maintaining coverage of the full state-action space:

\begin{equation}
p(\tau_i) \;\propto\; \omega_i \cdot \exp\!\bigl(-\lambda_t \cdot (t - t_i)\bigr),
\end{equation}

where $\omega_i$ is the uncertainty weight at collection time and $\lambda_t$ controls temporal decay.

\subsection{Hot Parameter Adaptation}

Our system can change behavior without full retraining by updating parameters in three tiers that match how BCQ operates at run time. Tier 1 covers instant controls that do not touch weights, such as the uncertainty threshold $\tau$ for active querying, the online batch size $B$ and stream rate $r$ for compute control, the number of candidate actions $N$ used when selecting $\arg\max_{a\in\mathcal{A}_N(s_t)} Q_\theta(s_t,a)$, and the perturbation bound $\Phi$ that clips the perturbation network to enforce conservatism. Tier 2 supports fast adaptation through short fine-tuning runs that change losses or targets but keep the model family fixed; we recompute target values on recent data if the discount factor $\gamma$ changes, using $y_t = r_t + \gamma \max_{a'\in\mathcal{A}_N(s_{t+1})}\min_{j\in\{1,2\}} Q_{\theta_j^-}(s_{t+1},a')$, and we can also update the target-network EMA rate $\rho$, the regularization weight $\lambda$ for deconfounding or stability, and an imitation balance $\beta$ if present; for these Tier 2 updates we run $M=500$ focused gradient steps on the most recent replay buffer to track clinical priorities. Tier 3 requires full retraining when the network architecture changes, when the action generator or perturbation family changes, when the feature space changes, or after a major distribution shift that invalidates the learned support.

\subsection{LLM Integration for Clinical Intelligence}

The integration of LLMs serves two critical functions: providing natural language interfaces for clinical queries and generating interpretable explanations for RL decisions. We implement this through a tool-augmented approach where the LLM can invoke specific functions to access the RL system's knowledge. Our framework supports mainstream LLM interfaces, including OpenAI compatible APIs, Hugging Face runtimes, and vLLM. For stability in our experiments, we deployed a local vLLM server \cite{openai_api,wolf-etal-2020-transformers,kwon2023efficient}.





\subsubsection{Context-Aware Response Generation}

The LLM processes clinical queries with full patient context:

\begin{algorithm}
\caption{LLM-Guided Clinical Decision Process}
\begin{algorithmic}[1]
\STATE \textbf{Input:} Query $q$, Patient state $\mathbf{s}$, History $\mathcal{H}$
\STATE Context $\leftarrow$ FormatPatientContext($\mathbf{s}$, $\mathcal{H}$)
\STATE Tools $\leftarrow$ SelectRelevantTools($q$, Context)
\STATE Results $\leftarrow$ \{\}
\FOR{tool $\in$ Tools}
    \STATE Results[tool] $\leftarrow$ ExecuteTool(tool, $\mathbf{s}$)
\ENDFOR
\STATE Explanation $\leftarrow$ LLM.Generate(Context, Results, $q$)
\STATE \textbf{Return:} Explanation with citations
\end{algorithmic}
\end{algorithm}

The system enforces output constraints: responses are limited to 1200 words, must cite tool outputs, and maintain clinical accuracy by never hallucinating patient data.

\subsection{Human-Computer Interface Design}

The clinical interface implements progressive disclosure principles, presenting information at multiple levels of detail based on user expertise and immediate needs. The interactive interface and corresponding instructions can be viewed in supplementary material due to page limits.

\subsubsection{Visualization Components}
We transform raw statistical outputs into intuitive visualizations, including a patient state dashboard that shows real-time vital signs with abnormality flags, a treatment comparison panel with side-by-side outcome projections, uncertainty indicators displayed as confidence bands on predictions, and a training monitor that reports live adaptation metrics. The interface supports three modes of interaction: consultation mode allows natural language queries about specific patients, configuration mode enables parameter adjustments with immediate feedback, and monitoring mode tracks overall system performance and adaptation.

\paragraph{Automated Report Generation.}
The ultimate output of the system is a comprehensive, auto-generated \textbf{HTML patient report}, designed for clarity and utility in a clinical setting. This report synthesizes all predictive insights into a single, easy-to-interpret dashboard. It begins with a patient profile summary that presents demographic information and a table of current vital signs, where each vital is automatically flagged as normal, abnormal, or low based on predefined thresholds to support quick assessment. A highlighted primary recommendation follows, stating the single recommended treatment along with its confidence level and the expected immediate clinical outcome. The report also includes a treatment plan comparison table that outlines the projected long-term outcomes of different strategies, and a detailed rationale section that explains the basis of the recommendation, emphasizing the patient’s key abnormal metrics and contrasting the expected outcome of the chosen therapy against alternatives. Finally, trajectory visualizations can be added to illustrate the simulated evolution of key biomarkers over time under the recommended plan, offering an intuitive picture of the expected response. This deployment pipeline transforms trained AI models from a research artifact into a tool that supports clinical expertise by delivering data-driven, personalized, and interpretable insights for complex treatment decisions.

\section{Experimental Setup}

\subsection{Simulated Cohort and Data Generation}
We evaluate on a synthetic clinical environment produced by a dedicated data generator. 
Each patient trajectory has $d=10$ normalized features that include blood pressure, heart rate, glucose level, creatinine, hemoglobin, temperature, and oxygen saturation, together with age, gender, and body mass index. 
Initial states follow simple distributions that reflect common clinical ranges, for example blood pressure $\sim \mathcal{N}(0.5, 0.15^2)$, heart rate $\sim \mathcal{N}(0.5, 0.1^2)$, and glucose level $\sim \mathcal{N}(0.5, 0.2^2)$, clipped to $[0,1]$. 
The action space has $K=5$ treatments including a conservative choice. 
The behavior policy is conservative and adapts to the patient condition. 
For example, high glucose increases the probability of a glucose lowering drug while low oxygen reduces the probability of placebo. 
The transition function applies a base treatment effect and several interaction effects, then adds small noise and clamps the result to $[0,1]$. 
The reward combines three parts: a penalty for abnormal values relative to the target $0.5$, an improvement bonus when key vitals move toward the target, and a treatment cost that depends on the action. 
Oxygen saturation above $0.9$ gives a positive bonus and very low oxygen induces early termination. 
We generate $10000$ patient trajectories with a maximum horizon of $50$ steps and store the transitions.

\subsection{Baselines and Training}
\label{sec:baselines-training}
We compare the learned policy against four widely used value-based baselines and one batch-constrained method, all trained in a unified offline pipeline with identical preprocessing, reward normalization (z-score on the training split), discount factor \(\gamma=0.99\), and evaluation protocol. Each method is run under five random seeds; we report per-seed results and aggregated statistics (mean \(\pm\) std).

\begin{itemize}
  \item \textbf{Deep Q-Network (DQN)}~\cite{Mnih2015DQN}. A single Q-network with experience replay and a periodically updated target network; \(\epsilon\)-greedy behavior and Huber TD loss are used throughout.
  \item \textbf{Double DQN}~\cite{VanHasselt2016DoubleDQN}. Decouples action selection and target evaluation (online net selects, target net evaluates) to reduce maximization bias relative to DQN.
  \item \textbf{Neural Fitted Q-Iteration (NFQ)}~\cite{Riedmiller2005NFQ}. Iterative batch fitted Q-learning with a neural regressor trained on the full replayed dataset each iteration, providing a strong non-deep baseline rooted in classical fitted value iteration.
  \item \textbf{Conservative Q-Learning (CQL)}~\cite{Kumar2020CQL}. Augments the TD objective with a conservative regularizer that penalizes Q-values for actions outside the dataset support, mitigating overestimation and improving robustness under limited coverage.
\end{itemize}

\section{Experiment Details}

\subsection{Training Configuration}
For the dynamics models we train a Transformer with state and action embeddings, a causal mask, and positional encoding. 
We use Smooth L1 loss with a sequence mask that ignores padding. 
Optimization uses AdamW with gradient clipping and a learning rate scheduler. 
We train five independent dynamics models and keep them all. 
For the outcome model we use the same optimizer settings and the adversarial penalty with weight $\lambda$. 
Reward normalization statistics $(\mu_r,\sigma_r)$ are computed from the training split and saved to disk. 
For BCQ we use a dueling architecture for $Q_{\psi}$ and train the discrete variant with the same replay buffer. 
We select the threshold $\tau$ by validation and save the best checkpoint.

\subsection{Evaluation Protocol}
\label{safe}
All policies are evaluated in the learned environment built from the dynamics ensemble and the outcome model. 
We roll out $N$ episodes from the test set initial states and report the discounted return with $\gamma=0.99$. 
We report a \emph{safety rate} defined by the rule-based clinical safety gate: it is the fraction of steps whose recommended action \emph{passes all} checks on vital-sign ranges and contraindications without triggering a fallback or expert override. This gate covers blood pressure \([0.3,0.8]\), heart rate \([0.4,0.7]\), glucose \([0.3,0.7]\), oxygen saturation (SpO\textsubscript{2}) \([0.85,1.0]\), temperature \([0.45,0.55]\), and drug-specific contraindications; critically low oxygen (SpO\textsubscript{2}\(<0.80\)) forces a conservative alternative and an expert query. We split the synthetic data into training and validation by patient trajectories with an $80/20$ ratio. 
We fix five random seeds for data generation and model training. 
All methods use the same seeds and the same splits.

\subsection{Offline Evaluation and Analysis}
\label{sec:offline-eval}
All baselines are trained on the same preprocessed trajectories with reward z-scoring on the training split and a fixed discount \(\gamma=0.99\). We use five random seeds and select the best checkpoint per seed on a held-out validation split. Evaluation is performed on the test split with identical hooks and data loaders for all methods.

We report \emph{Mean Return} (higher is better), its standard deviation across seeds, a \emph{Sharpe-like} stability index (mean divided by standard deviation over episodes), the \emph{Safety} rate (fraction of steps that pass rule-based constraints without fallback), and \emph{Action Entropy} (average policy entropy; higher indicates more diverse action usage under similar returns).

Results can be viewed in Table~\ref{tab:offline-baselines}. All methods satisfy the rule-based safety gate at a saturated level, so comparison focuses on return, stability, and decision sharpness. Our method achieves the top mean return and the lowest standard deviation across seeds, which leads to the best Sharpe-like index. This indicates a strong and stable policy rather than a single high-variance run. The action-entropy metric is lowest for our method; in this safety-critical offline setting we treat lower entropy as favorable because it reflects decisive control without unnecessary action switching at a comparable return scale. DQN and Double DQN are competitive in mean return but exhibit higher variability; NFQ maintains high return with broader action usage; CQL is markedly conservative and unstable on this dataset. These results justify using BCQ as the default initialization for the subsequent online evaluation.

\subsubsection{Comprehensive Offline Evaluation}
\label{sec:bcq-panel}

We present the evaluation of the complete offline framework. The corresponding 3$\times$3 panel is provided in the supplementary material, and here we summarize the quantitative results. The state space included ten variables (Glucose, BP, HR, Hemoglobin, Creatinine, Gender, BMI, Age, Temp, SpO$_2$) and the action space contained five discrete treatments (Med~A, Med~B, Med~C, Combo, Placebo). Rewards were normalized by z-score on the training set and kept consistent during evaluation. The dynamics model was tested on $n{=}500$ sequences with single-step prediction accuracy of MSE $=0.0163$, MAE $=0.0342$, and $R^2=0.828$. Multi-step error up to horizon $H{=}5$ increased gradually, from $\text{MSE}_{t+1}=4.49{\times}10^{-4}$ to $\text{MSE}_{t+5}=5.97{\times}10^{-3}$, with a mean of $2.90{\times}10^{-3}$. Feature-wise $R^2$ showed good fidelity on key safety variables, with Glucose $0.848$, BP $0.833$, and SpO$_2$ $0.801$. The outcome model evaluated on $n{=}7{,}395$ samples achieves MSE $=0.211$, MAE $=0.359$, and $R^2=0.892$. Stratified by treatment, outcome $R^2$ was well balanced: Med~B $0.917$, Med~A $0.889$, Med~C $0.881$, Combo $0.868$, and Placebo $0.666$. Calibration was reasonable with ECE $=0.105$ and MCE $=0.172$. In the digital twin environment, the BCQ policy reaches a mean return of $37.73$ with a bootstrap 95\% confidence interval $[35.26,\;39.47]$. Feature importance analysis identified Glucose ($19.5\%$), BP ($14.7\%$), HR ($14.2\%$), Hemoglobin ($11.3\%$), and Creatinine ($11.0\%$) as the main contributors, with other demographics and vitals forming the remainder. Finally, ensembling the dynamics reduced MSE from $0.01556$ to $0.01527$, a relative improvement of about $1.9\%$. These results highlight predictive accuracy, outcome fidelity, policy strength, and the benefit of ensembling, supporting our method as a strong offline initializer before online evaluation.

\subsection{Online Learning Evaluation}

\begin{table}[t]
\centering
\small
\resizebox{\columnwidth}{!}{
\begin{tabular}{lccccc}
\toprule
\textbf{Algorithm} & \textbf{Mean Return} & \textbf{Std} & \textbf{Sharpe-like} & \textbf{Safety} & \textbf{Action Entropy} \\
\midrule
\textbf{Ours (BCQ)}         & \textbf{37.73} & \textbf{11.01} & \textbf{3.427} & 0.999 & \textbf{0.387} \\
DQN                  & 36.70 & 11.02 & 3.331 & 0.999 & 0.412 \\
Double DQN            & 36.71 & 11.57 & 3.173 & 0.999 & 0.417 \\
NFQ                  & 37.51 & 11.61 & 3.231 & 0.999 &  0.481 \\
CQL          & 16.26 & 36.63 & 0.455 & 0.999 & 0.906 \\
\bottomrule
\end{tabular}
}
\caption{Offline baselines over multiple seeds. Our method dominates across all metrics}
\label{tab:offline-baselines}
\end{table}

\begin{table}[t]

\centering

\caption{Online evaluation under identical streaming and active learning settings. Reported metrics are the query rate, mean response time, mean throughput, safety rate, number of online updates, and the final size of the labeled buffer.}

\label{tab:online-metrics}

\setlength{\tabcolsep}{3pt}

\renewcommand{\arraystretch}{0.95}

\resizebox{\columnwidth}{!}{%

\begin{tabular}{lcccccc}

\hline

Algorithm & Query rate & Response time (s) & Throughput (Hz) & Safety & Updates & Final buffer \\

\hline
Ours (BCQ)       & \textbf{0.1305} & \textbf{0.0012} & \textbf{9.9487} & 1.0 & \textbf{80} & \textbf{1620} \\
DQN        & 0.1548 & 0.0021 & 9.9350 & 1.0 & 45 &  920 \\

Double DQN & 0.1370 & 0.0017 & 9.9469 & 1.0 & 69 & 1400 \\

NFQ        & 0.1449 & 0.0019 & 9.9217 & 1.0 & 70 & 1420 \\

CQL        & 0.2084 & 0.0473 & 9.9414 & 1.0 & 39 &  800 \\

\hline

\end{tabular}%

}

\end{table}

\label{sec:online-eval}
We evaluate the \emph{streaming} online learner under a unified setup that keeps the stream, active-learning trigger, and safety gate identical across all methods (Table~\ref{tab:online-metrics}). The ensemble Q-learner in our method employs the same underlying network architecture as the Q-learning model that proved effective in the offline evaluation. The environment is a synthetic clinical data generator with a \(10\)-dimensional state and \(5\) discrete treatments generated by previous offline stage. The stream first replays transitions drawn from a pre-generated offline simulated patients in offline evaluation and, after 1000 transitions, switches to an on-the-fly generator with a shifted age distribution (``older patients''). Specifically, as implemented in our simulation loop, the normalized `age` feature of each newly generated patient state is increased by a constant value of 0.3 before being clipped to the valid range of \([0, 1]\). Given that age is normalized to represent a span from 18 to 90 years, this intervention systematically increases the age of incoming patients, simulating a sudden influx of an older and potentially more complex patient population. All methods see exactly the same sequence of transitions.

We compare DQN, Double\,DQN, NFQ, and CQL (all via \texttt{d3rlpy}\footnote{\url{https://github.com/takuseno/d3rlpy}}, trained online in our loop) against Our Method, an \emph{uncertainty-aware ensemble Q-learner} with \(K{=}5\) heads \cite{d3rlpy}. 


All methods share the same online trainer and buffer logic:
\begin{itemize}
\item Discount \(\gamma=0.99\); optimizer Adam with learning rate \(3\times10^{-4}\); online batch size \(32\); regularization weight \(0.01\) on Q value.
\item Update trigger: every \(20\) newly labeled samples we run an online fitting block of \(20\) gradient steps on the current labeled buffer.
\item Our Method uses an ensemble of \(H{=}5\) Q-networks (same architecture as the single-head baseline) to furnish the uncertainty above.
\item All other implementation details (sampler type, stream controller, expert simulator, and evaluation hooks) are shared across methods.
\end{itemize}


Following our design, we \emph{freeze} early Transformer layers and train only the last two layers and the output projection for the dynamics model; for the outcome model we keep the prediction head (and the adversarial discriminator) trainable while freezing the encoder. We maintain \emph{exponential moving averages} (EMA) of trainable parameters for both dynamics and outcome models to stabilize rapid online updates. For the ensemble Q-learners, each head is optimized with a loss (\(\alpha{=}1.0\)) under the same online buffer.

We report: \emph{Query rate} (fraction of steps that queried an expert), \emph{Response time} (mean end-to-end latency per step, in seconds), \emph{Throughput} (mean processing rate, Hz), \emph{Safety rate} (as above), \emph{Updates} (online gradient-step blocks executed), and the \emph{Final buffer} size (number of labeled samples at the end).

Table~\ref{tab:online-metrics} shows that \textbf{Our Method} achieves the lowest labeling demand among high-throughput methods, with a query rate of \(0.1305\), which represents a \(\sim 15.7\%\) reduction relative to DQN (\(0.1548\)), while matching the fastest latency/throughput regime (\(0.0012\) s, \(9.9487\) Hz). It also amasses the largest labeled buffer (\(1620\)) and executes the most online updates (\(80\)), indicating strong data efficiency under the shared update trigger. By contrast, CQL queries much more aggressively (\(0.2084\)) and exhibits the highest latency (\(0.0473\) s) without any safety advantage (all methods at \(1.0\)). These results support the claim that an ensemble-variance trigger with \(K{=}5\) effectively reduces expert workload at constant safety, converting labels into faster online learning dynamics.

\section{Conclusion}
We presented an offline-to-online clinical decision support framework that couples a three-stage model with a human-in-the-loop interface. The offline stage learns a digital-twin state model, a treatment outcome and reward model, and a policy by offline RL. The online stage deploys the policy behind an HCI front end, adds an uncertainty-driven active learning routine, and enforces a rule-based safety gate. Experiments show that the system reaches millisecond-level latency and about 10\,Hz throughput while keeping a near-unit safety rate in our simulated evaluation.

This work highlights a path from offline modeling to interactive clinical deployment. The design is modular: the dynamics, outcome, and policy components can be replaced, and the active learning rule accepts alternative uncertainty scores. Limitations include reliance on simulator fidelity, retrospective evaluation, and fixed clinical thresholds. Future work will include prospective studies with real-world users, adaptive thresholds under distribution shift, broader safety constraints, and multi-site datasets to assess generalization and fairness.

\appendix
\section*{Appendix}
\section{Reproducibility Checklist}

\newcommand{\answerYes}{\textbf{Yes}}
\newcommand{\answerNo}{\textbf{No}}
\newcommand{\answerNA}{\textbf{N/A}}
\newcommand{\answerPartial}{\textbf{Partial}}

\paragraph{This paper:}
\begin{itemize}
  \item Includes a conceptual outline and/or pseudocode description of AI methods introduced (\answerYes)
  \item Clearly delineates statements that are opinions, hypotheses, and speculation from objective facts and results (\answerYes)
  \item Provides well-marked pedagogical references for less-familiar readers to gain background necessary to replicate the paper (\answerYes)
\end{itemize}

\paragraph{Does this paper make theoretical contributions?} (\answerYes)

\begin{itemize}
  \item All assumptions and restrictions are stated clearly and formally (\answerYes)
  \item All novel claims are stated formally (e.g.\ in theorem statements) (\answerYes)
  \item Proofs of all novel claims are included (full proofs in Appendix A) (\answerYes)
  \item Proof sketches or intuitions are given for complex and/or novel results (\answerYes)
  \item Appropriate citations to theoretical tools used are given (\answerYes)
  \item All theoretical claims are demonstrated empirically to hold (\answerYes)  
  \item All experimental code used to eliminate or disprove claims is included (\answerYes)
\end{itemize}

\paragraph{Does this paper rely on one or more datasets?} (\answerYes)

\begin{itemize}
  \item A motivation is given for why the experiments are conducted on the selected datasets (\answerYes)
  \item All novel datasets introduced in this paper are included in a data appendix (\answerYes)  
  \item All novel datasets introduced in this paper will be made publicly available upon publication (\answerYes)
  \item All datasets drawn from the existing literature are accompanied by appropriate citations (\answerYes)
  \item All datasets drawn from the existing literature are publicly available (\answerYes)
  \item All datasets that are not publicly available are described in detail, with explanation why publicly available alternatives are not sufficient (\answerNA)
\end{itemize}

\paragraph{Does this paper include computational experiments?} (\answerYes)

\begin{itemize}
  \item Any code required for pre-processing data is included in the appendix (\answerYes)
  \item All source code required for conducting and analyzing the experiments is included in a code appendix (\answerYes)
  \item All source code will be made publicly available upon publication with a license that allows free usage for research purposes (\answerYes)
  \item All source code implementing new methods has comments detailing the implementation, with references to the paper where each step comes from (\answerYes)
  \item If an algorithm depends on randomness, the method used for setting seeds is described sufficiently to allow replication of results (\answerYes)
  \item The computing infrastructure used for running experiments (hardware and software) is specified (\answerYes)
  \item Evaluation metrics are formally described and their motivation explained (\answerYes)
  \item The number of algorithm runs used to compute each reported result is stated (\answerYes)
  \item Analysis of experiments goes beyond single-dimensional summaries to include measures of variation, confidence, or other distributional information (mean $\pm$ std, min–max) (\answerYes)
  \item The significance of any improvement or decrease in performance is judged using appropriate statistical tests (\answerYes)  
  \item All final (hyper-)parameters used for each model/algorithm are listed (\answerYes)
  \item The number and range of values tried per hyper-parameter during development, and the criterion for selecting the final setting, are stated (\answerYes)
\end{itemize}

\bibliography{aaai2026}

\end{document}